\documentclass[paper]{ieice}
\usepackage[fleqn]{amsmath}
\usepackage{newtxtext}
\usepackage[varg]{newtxmath}
\usepackage{booktabs}
\usepackage{graphicx}
\usepackage{subfigure}
\usepackage{array}%需要该宏包

\field{}
\title {Security Consideration For Deep Learning-Based Image Forensics}
\authorlist{%
 \authorentry[w$\_$zhao@bjtu.edu.cn]{Wei Zhao}{n}{affiliate label}\MembershipNumber{}
 \authorentry[ppyang@bjtu.edu.cn]{Pengpeng Yang}{n}{affiliate label}\MembershipNumber{}
 \authorentry[rrni@bjtu.edu.cn]{Rongrong Ni}{m}{affiliate label}\MembershipNumber{12356}
 \authorentry[yzhao@bjtu.edu.cn]{Yao Zhao}{n}{affiliate label}\MembershipNumber{}
 \authorentry[hrwu@bjtu.edu.cn]{Haorui Wu}{n}{affiliate label}\MembershipNumber{}
}
\affiliate[affiliate label]{The author is with the
}
\received{2015}{1}{1}
\revised{2015}{1}{1}

\begin{document}
\maketitle
\begin{summary}
Recently, image forensics community has paid attention to the research on the design of effective algorithms based on deep learning technology. And facts proved that combining the domain knowledge of image forensics and deep learning would achieve more robust and better performance than the traditional schemes. Instead of improving algorithm performance, in this paper, the safety of deep learning based methods in the field of image forensics is taken into account. To the best of our knowledge, this is a first work focusing on this topic. Specifically, we experimentally find that the method using deep learning would fail when adding the slight noise into the images (adversarial images). Furthermore, two kinds of strategys are proposed to enforce security of deep learning-based methods. Firstly, a penalty term to the loss function is added, which is the 2-norm of the gradient of the loss with respect to the input images, and then an novel training method is adopt to train the model by fusing the normal and adversarial images.  Experimental results show that the proposed algorithm can achieve good performance even in the case of adversarial images and provide a safety consideration for deep learning-based image forensics.

\end{summary}

\section{Introduction}
With the rapid development of network technology and the popularity of digital cameras, images have become a important information carrier. At the same time, with widespread use of image editing tools, the authenticity and integrity of images are greatly challenged. Some people use these tools to tamper with images maliciously, which can make a series of negative impacts on the society, especially in the field of news, military, business and judicial expertise. For example, some attackers use tampered images for military or political purpose, which can cause a crisis of confidence. Therefore, digital image forensics technologies arise at the historic moment for verifying the authenticity of images.

In the past decade, a number of algorithms have been proposed in image forensic community. In particular, methods based on deep learning play an important role due to its excellent performance. Yang $et.al$[1] proposed the laplacian convolutional neural networks to detect recaptured images. Kang $et.al$[2] combined convolutional neural networks(CNN) and MFR to deal with median filtering forensics and Tang $et.al$[3] improved its performance using MFNet. For source camera identification, the schemes using deep learning have also been proposed in these works[4-7]. Barni $et.al$[8] used three kinds of CNN architecture: CNN in the pixel domain, CNN in noise domain, and CNN embedding DCT histograms, to detect double JPEG compressed images. In addition, Bayar $et.al$[9] presented a deep learning approach for universal image manipulation detection. Of special interest is that the above algorithms have better performance than the traditional methods.

Despite superior performance of deep learning-based methods, there lies a security threat: adversarial examples. According to the reports[10-15], a deep learning model can get an error output with high confidence for the input data with added slightly noise(adversarial examples). For this phenomenon, there has been no comments so far from the image forensics community. However, the safety of deep-learning methods should be be considered, otherwise it will cause serious consequences. For example, recaptured image forensic can effectively resist the face presentation attack. If the deep learning-based recaptured image detection algorithms is easily misleaded by adversarial examples, personal and property safety protected by face recognition system will be threatened.

In this paper, focusing on the safety of deep learning-based image forensic algorithms, we first experimentally verify its vulnerability by adding noise to the clean images. Specially, two different methods of generating adversarial examples are referred, that is, adding perturbation in the direction of the biggest change of the loss function or in the direction of being classified as least-likely class. It should be noted that, in order to simplify our discussion, the recaptured image forensic scheme based on laplacian convolutional neural network, as a typical deep learning-based image forensic scheme, is taken into account in this work. Then, in order to resist against attack of adversarial examples, a penalty term to the loss function is added which is the 2-norm of the gradient of the loss with respect to the input images, and an novel training method is adopt to train the model by fusing the normal and adversarial images. Experimental results show that the proposed method has the advantages of safety and achieve great performance in the cases of different methods of adding noise with random strength.

The rest of the paper is organized as follows. In Section 2, the structure of laplacian convolutional neural network is briefly introduced. In Section 3, the vulnerability of deep learning-based method to adversarial examples is explained and  a safe CNN-based scheme to resist against attack of noise is proposed. The experiment results are conducted in Section 4, and conclusions are drawn in Section 5.
\section{Previous Work}
In this section we briefly introduce a deep learning-based recaptured image forensic algorithm [1] which achieved state-of-the-art performance. The  architecture of the work include two parts: single enhancement layer and general convolutional neural networks structure. In the single enhancement layer, laplacian filter was used to enhance the signal.
\mathindent=25mm
 \begin{equation}\label{1}
   LF=\left[
     \begin{array}{ccc}
       0, & -1, & 0 \\
       -1, & 4,& -1 \\
       0, & -1, & 0 \\
     \end{array}
   \right]
 \end{equation}
\mathindent=7mm

In the second part, the filtered images are fed into the first convolution unit which includes: a convolution layer, a batch normalization layer, relu function and an average pooling layer. The later four convolution units have the same composition as the first unit with only one difference that the pooling layer in the fifth unit is replaced by global average pooling. Finally, a full connection layer is used as a classification layer. For more details, the kernels size of convolution layer is $3\times3$ with 1 step size, and the kernels size of pooling layer is $5\times5$ with 2 step size.

\section{Proposed Method}

Although image forensics schemes based on deep learning have achieved exceptional performance, there still be a serious drawback. That is, images will be likely to be misclassified if some noise is added onto the images in a particular way. In this section, firstly, the vulnerability of the deep learning-based image forensics algorithm is experimentally verified. Then, two kinds of strategies which can resist noise attack effectively to increase security of the deep learning-based methods are present.

\subsection{Vulnerability Analysis of Deep Learning-Based Image Forensics}

\begin{figure}
  \centering
  \includegraphics[width=5cm]{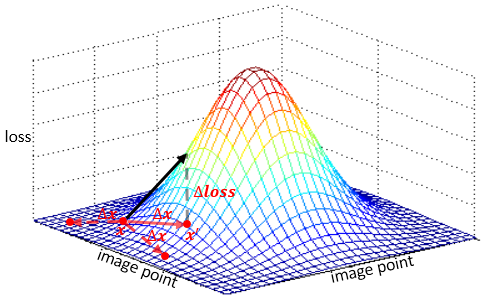}\\
  \caption{The explain of the fast change direction}\label{1}
\end{figure}

In the Figure. 1, suppose $l(\cdot)$ is the loss function and $x$ is an sample point that can be classified correctly. For a well-trained model, $l(x)$ is a relatively small values. If a slight perturbation $\Delta x$ is added onto $x$ along a particular direction, the $l(x+\Delta x)$ will become a larger value, which means the classification result will be changed with high possibility.

According to the above idea, two specific methods of adding noise are used. For simplicity, the notations are as follows: \begin{itemize}
  \item $x_{clean}$ - A normal image example, the pixels values of which are integer numbers in the range [0,255].
  \item $y_{label}$ - True class for an image
  \item $x_{adv}$ - An adversarial example generated by some method.
  \item $L(x,y)$ - Loss function that used to train a neural network.
  \item $sign(x)$ - Sign function. $sign(x)=\left\{
            \begin{array}{ll}
              1, & \hbox{$x<0$} \\
              0, & \hbox{$x=0$} \\
              -1, & \hbox{$x>0$}
            \end{array}
          \right.$

  \item $clip(x,max,min)$ -  Clip function.

$clip(x,max,min)=\left\{
            \begin{array}{ll}
              max, & \hbox{$x>max$} \\
              x, & \hbox{$min<x<max$} \\
              min, & \hbox{$x<min$}
            \end{array}
          \right.$
\end{itemize}

\begin{figure}
\centering
  \subfigure[$y_{true}$: rec \newline \centerline{$y_{pred}$: org} \newline \centerline{$prob=100\%$}] {\includegraphics[width=1.8cm]{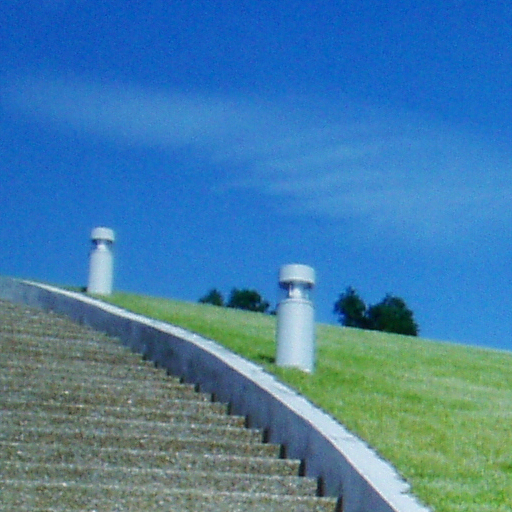}}\hspace{.3cm}
  \subfigure[$y_{true}$: rec \newline \centerline{$y_{pred}$: org} \newline \centerline{$prob=100\%$}] {\includegraphics[width=1.8cm]{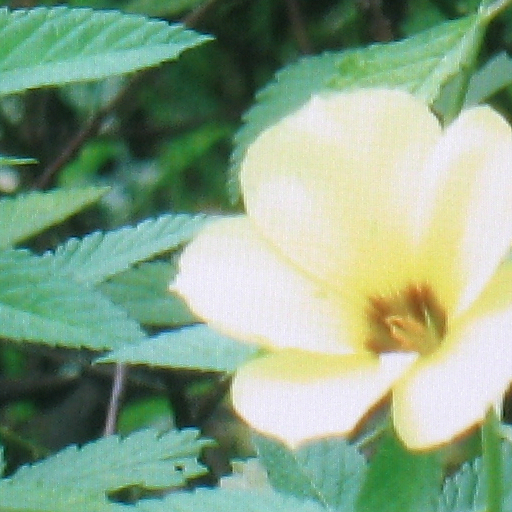}}\hspace{.3cm}
  \subfigure[$y_{true}$: org \newline \centerline{$y_{pred}$: rec} \newline \centerline{$prob=100\%$}] {\includegraphics[width=1.8cm]{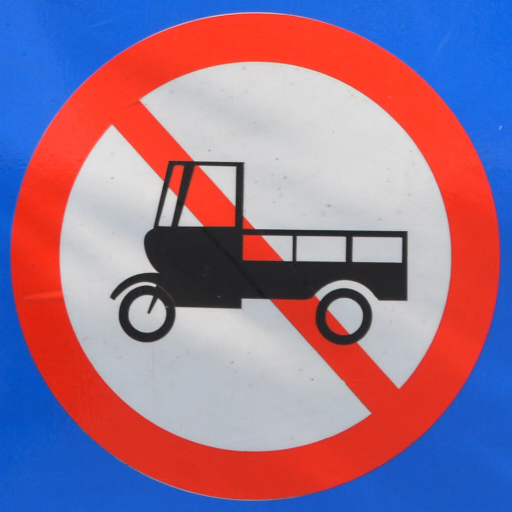}}\hspace{.3cm}
  \subfigure[$y_{true}$: org \newline \centerline{$y_{pred}$: rec} \newline \centerline{$prob=99.9\%$}] {\includegraphics[width=1.8cm]{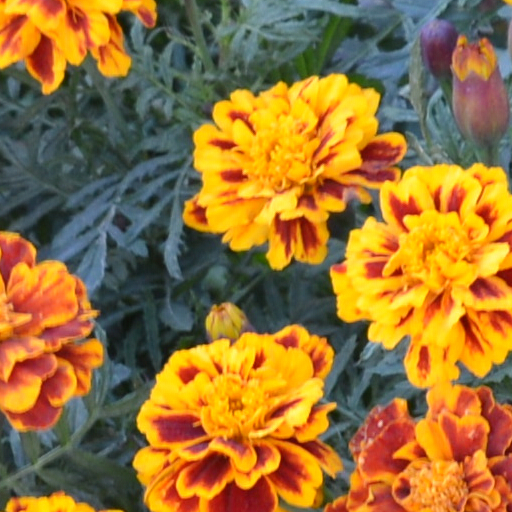}}\hspace{.3cm}

  \subfigure[$y_{true}$: rec \newline \centerline{$y_{pred}$: rec} \newline \centerline{$prob=100\%$}] {\includegraphics[width=1.8cm]{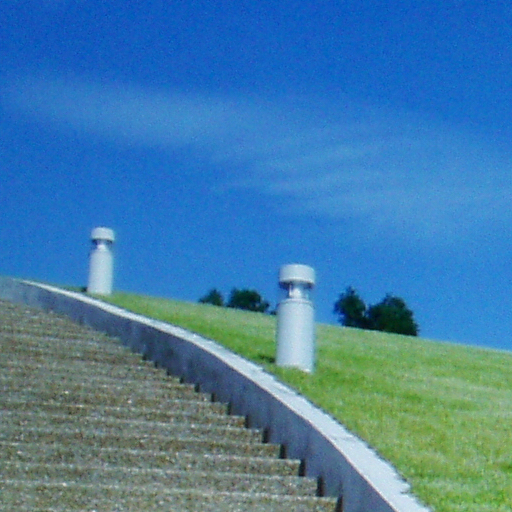}}\hspace{.3cm}
  \subfigure[$y_{true}$: rec \newline \centerline{$y_{pred}$: rec} \newline \centerline{$prob=100\%$}] {\includegraphics[width=1.8cm]{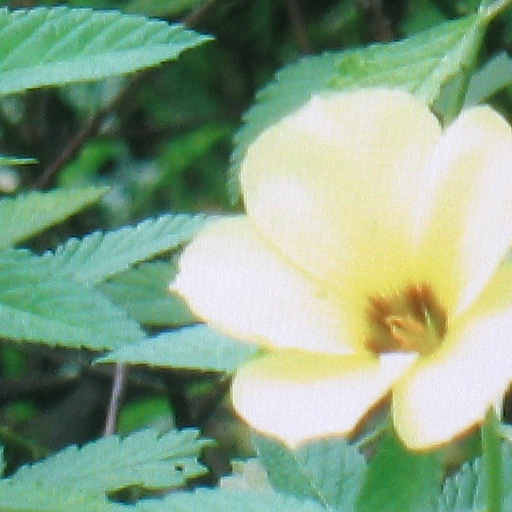}}\hspace{.3cm}
  \subfigure[$y_{true}$: org \newline \centerline{$y_{pred}$: org} \newline \centerline{$prob=100\%$}] {\includegraphics[width=1.8cm]{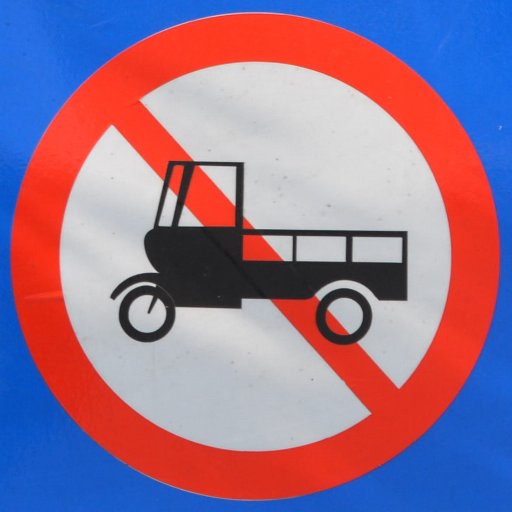}}\hspace{.3cm}
  \subfigure[$y_{true}$: org \newline \centerline{$y_{pred}$: org} \newline \centerline{$prob=100\%$}] {\includegraphics[width=1.8cm]{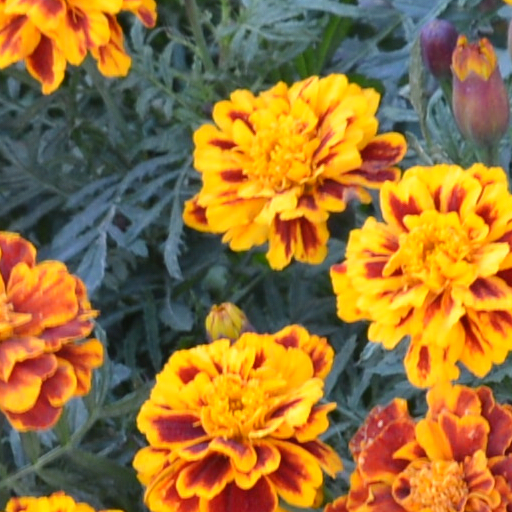}}\hspace{.3cm}
  \caption{The adversarial examples and their detection results which is generated by FGSM with $\epsilon=1$ using Yang's and proposed methods, respectively. (a) (b) (c) (d) is generated and detected with Yang's model. (e) (f) (g) (h) is generated and detected with proposed model.}\label{3}
\end{figure}

Fast Gradient Sign Method (FGSM) is to find the direction by making the partial derivatives of the loss by the input image $x_{clean}$. By adding noise on the clean image examples along this direction, the loss will be changed most and the probability of making a misclassification is the highest.
\mathindent=0mm
\begin{equation}\label{2}
  x_{adv}\!=\!clip\{x_{clean}+\epsilon sign(\frac{\partial L(x_{clean},y_{label})}{\partial x_{clean}}),0,255\}
\end{equation}
\mathindent=7mm

Unlike the FGSM method using the true label to compute the direction, Least-Likely Class Method (LLCM) use the predicted label of the well-trained model to compute the direction and hoping to make the noised images be classified as the least likely class.
\mathindent=25mm
\begin{equation}\label{3}
  y_{min}=\arg _{y} \min \{p(y|x)\}
\end{equation}
\mathindent=0mm
\begin{equation}\label{4}
  x_{adv}=clip\{x_{clean}-\epsilon sign(\frac{\partial L(x_{clean},y_{min})}{\partial x_{clean}}),0,255\}
\end{equation}
\mathindent=7mm

There are some adversarial examples which are generated using FGSM with $\epsilon=1$ in Figure. 2, and the upper part is adversarial examples generated by Yang's model and corresponding detection results. And $y_{true}$ is the true label of each image,  $y_{pred}$ is the predicted label, and $prob$ is the probability that the image is classified as predicted label. One can be seen that the attack effect is so obvious that slight noise can lead to incorrect classification with high probability.

\subsection{Secure Image Forensics Using Deep Learning}

\begin{figure}
  \centering
  \includegraphics[width=8cm]{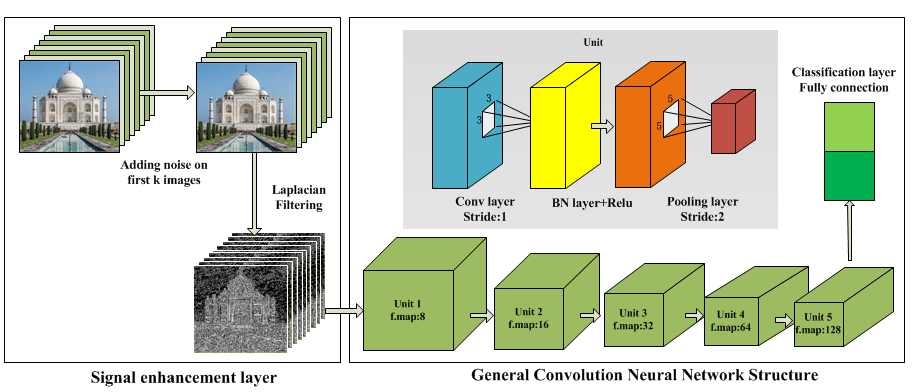}\\
  \caption{The architecture of Convolutional Neural Networks used in this work}\label{1}
\end{figure}

To counter against the threat of adversarial examples, we proposed a secure deep learning method for image forensics. The architecture of our CNN model is showed in Figure. 3. The main idea is to make the loss relatively smooth on the input images and their neighboring regions. Firstly, a penalty term is added on the loss function which is the 2-norm of the gradient of loss with respect to clean images. So the loss function is shown as equation (5).

\mathindent=20mm
\begin{equation}\label{6}
  L(x,y)= J(x,y)+\lambda \|\nabla_{x}J(x,y)\|_{2}
\end{equation}
where, $\lambda$ is the weight coefficient and $J$ is binary cross-entropy:
\mathindent=20mm
\begin{equation}\label{6}
  J(x,y) = -\sum_{i=1}^{2}log(\frac{e^{z_{i}}}{\sum_{k=1}^{2}e^{z_{k}}})
\end{equation}
 where $z$ is output of the last layer of the model.

 This term can constraint the loss to be relatively smooth so that not change too much when the input changes slightly, which equals predictions will not change significantly.

 Moreover, a novel training strategy is applied by fusing adversarial examples and clean images together as training set. Its process is as shown in Figure. 4, for every mini-batch $B = \{x^{1},x^{2},x^{3},\cdots x^{m}\}$ including $m$ clean images, the first $k$ $(k\leq m)$ images are took out and added noise. Then the rest clean images and the noised examples are merged to a new mini-batch $B^{'}= \{x_{noise}^{1},x_{noise}^{2},\cdots x_{noise}^{k},x_{clean}^{k+1},x_{clean}^{k+2},\cdots x_{clean}^{m}\}$. And when computing the loss of the new mini-batch, different weights to the clean image loss $L_{clean}$ and adversarial image loss $L_{noise}$ are applied. The loss of one batch can be computed:
\mathindent=4mm
\begin{equation}\label{6}
  Loss= \Sigma_{i\in x_{clean}}L(x_{i},y_{i})+\alpha\Sigma_{i\in x_{noise}}L(x_{i},y_{i})
\end{equation}

where, $\alpha$ is weight parameter. In our experiment, we empirically set $m$=64, $k$=32, and $\alpha=0.2$.

Noted that for each step, only one method - FGSM is used to generate adversarial examples with the loss of current training phase. In order to ensure that proposed method can be against various degrees of attack, adversarial examples are generated using different strength $\epsilon$ which is chosen in a uniform distribution. By doing so, the model will converge to a point where the loss is a relatively small value whether the input is clean images or noised images. This equals that the loss is relatively smooth in local neighborhoods of training points.

\begin{figure}
  \centering
  \includegraphics[width=7cm]{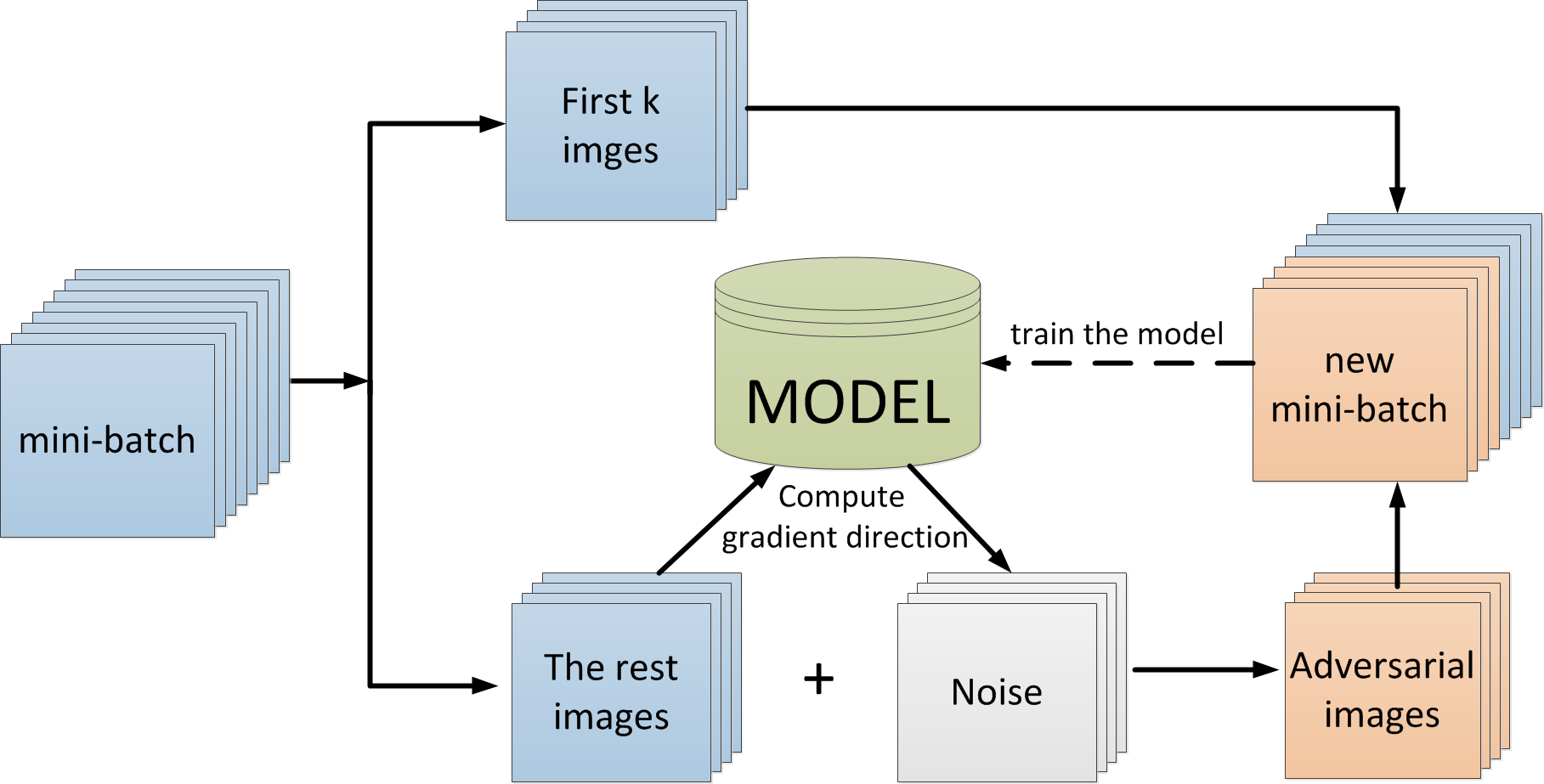}\\
  \caption{The process of training the model}\label{1}
\end{figure}

The second row in the Figure. 2 shows the adversarial examples generated by proposed model using FGSM with $\epsilon=1$ and corresponding detection results. It can be observed that the proposed method can resist against the adversarial examples' attack effectively and make a correct classification with high confidence.

\section{Experiments}

In this section, firstly, the vulnerability of the original deep learning scheme are evaluated by adding noise on the clean image with two different methods. And then the safety performance of proposed deep learning method to adversarial examples are checked.

The images used in our experiment are exactly the same as the work [1]. Four groups of datasets with size of $64\times64$, $128\times128$, $256\times256$ and $512\times512$ are used. Each group has 10000 original images and 10000 recaptured images. Then every group is divided into training set, validation set and test set by percent 40/10/50. the batch size is set as 64 and iteration number is set as 100k. The initial learning rate is 0.0001 and decays to 0.9 times every 10k iteration. The noise strength $\epsilon$ is chosen by a series of experiments. Some attacked images are given in Figure. 5, (a) and (e) are clean images, and (b) (c) (d) are adversarial images generated by FGSM with noise strength $\epsilon=5$, $\epsilon=10$, $\epsilon=15$, respectively, (d) (e) (f) are adversarial images generated by LLCM.  It can be found that the distortion of the images become ever more obvious with the increase of the noise intensity. While, slight noise ($\epsilon\leq 5$) is hard to be observed. So noise strength $\epsilon$ is chosen in the range of $[1.0, 5.0]$.

\subsection{Experiment 1}
In order to evaluate the vulnerability of Yang's scheme to adversarial examples. We first train laplacian convolutional neural network model by Yang' algorithm and generate adversarial examples using two methods on our test dataset which include 10000 images in one group of each size. The detection accuracy for clean images is shown in Table 1 and the classification accuracy of adversarial images is described in the Table 2.

One can be seen from the Table 2 that these two methods are both effective in attacking the model trained by deep learning based methods. FSGM is slightly better than LLCM no matter what size of the images. And there is a tendency that the bigger the image sizes are, the smaller gap between FSGM and LLCM. This is because that LLCM use predicted result to compute the gradient direction instead of the true label, so better attack effect can be made when the model can give more accurate result. And that is be proved in n the work[1] that larger images can get higher classification accuracy.

Another phenomenon need to be explained that the attack effect is not always raised with the increase in noise strength. We guess the reason is that the optimum point is crossed when a relatively large noise strength is used.

\begin{figure}
\centering
  \subfigure[Clean($\epsilon$=0)] {\includegraphics[width=1.8cm]{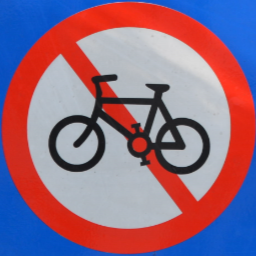}}\hspace{.3cm}
  \subfigure[ FGSM ($\epsilon$=5)] {\includegraphics[width=1.8cm]{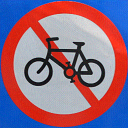}}\hspace{.3cm}
  \subfigure[ FGSM ($\epsilon$=10)] {\includegraphics[width=1.8cm]{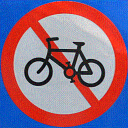}}\hspace{.3cm}
  \subfigure[ FGSM ($\epsilon$=15)] {\includegraphics[width=1.8cm]{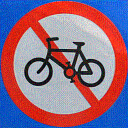}}\hspace{.3cm}

  \subfigure[Clean($\epsilon$=0)] {\includegraphics[width=1.8cm]{clean.png}}\hspace{.3cm}
  \subfigure[LLCM  ($\epsilon$=5)] {\includegraphics[width=1.8cm]{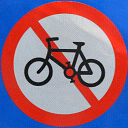}}\hspace{.3cm}
  \subfigure[LLCM  ($\epsilon$=10)] {\includegraphics[width=1.8cm]{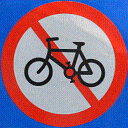}}\hspace{.3cm}
  \subfigure[LLCM  ($\epsilon$=15)] {\includegraphics[width=1.8cm]{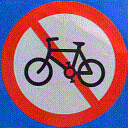}}\hspace{.3cm}
  \caption{visual effect of images based on different attack strength and different attack methods}\label{3}
\end{figure}

\subsection{Experiment 2}
In this part, the attack effect of adversarial images to proposed model are verified. For clean image, proposed method does not affect the accuracy much. In Table 1, the accuracy of clean images is displayed and it just has a slight and acceptable decline compared with Yang's.

In the Table 3, detection accuracy of adversarial images generated with two methods in different sizes are presented. On can be seen that our method have a significant effect in against of adversarial examples.

Note that we only inject FGSM adversarial examples into our training set, so proposed model can be against the FGSM noised examples very well. While LLCM adversarial examples are generated in a different way, so the resistance is not as great as FGSM.

\makeatletter
\def\hlinew#1{%
  \noalign{\ifnum0=`}\fi\hrule \@height #1 \futurelet
   \reserved@a\@xhline}
\makeatother
\begin{table}[htb!]
  \centering
  \renewcommand\arraystretch{1}
  \caption{The accuracy of clean images in different sizes }\label{2}
  \begin{tabular}{c|c|c|c|c}
  \hlinew{1pt}
  image size&$64\times64$ &$128\times128$ &$256\times256$ &$512\times512$ \\
  \hlinew{1pt}
  Yang &96.7\% &98.2\% &99.1\%  &99.4\%\\
  \hlinew{0.5pt}
  Prop &96.6\%&98.1\% &98.5\% &99.0\%\\
  \hlinew{1pt}
  \end{tabular}
\end{table}

\begin{table}[htb!]
  \centering
  \renewcommand\arraystretch{1}
  \caption{Detection Accuracy of Adversarial Images for Yang's Method }\label{2}
  \begin{tabular}{c|c|c|c|c|c|c}
  \hlinew{1pt}
  size &method&$\epsilon=1$ &$\epsilon=2$ &$\epsilon=3$ &$\epsilon=4$&$\epsilon=5$ \\
  \hlinew{1pt}
  $64\times64$  &FGSM &2.8\% &0.9\% &0.5\%  &0.6\%   &2.6\%\\
                &TGSM &7.5\% &5.7\% &5.2\%  &5.3\%   &6.9\%\\
  \hlinew{0.5pt}
  $128\times128$&FGSM &1.1\% &0.2\% &1.4\%  &4.9\%   &6.5\%\\
                &TGSM &2.9\% &2.0\% &3.1\%  &6.1\%   &7.5\%\\
  \hlinew{0.5pt}
  $256\times256$&FGSM &2.6\% &3.6\% &5.4\%  &7.5\%   &9.9\%\\
                &TGSM &3.9\% &4.9\% &6.6\%  &8.6\%   &11.1\%\\
  \hlinew{0.5pt}
  $512\times512$&FGSM &3.3\% &1.6\% &0.9\%  &0.7\%   &1.0\%\\
                &TGSM &4.6\% &2.8\% &2.2\%  &1.9\%   &2.1\%\\
  \hlinew{1pt}
  \end{tabular}
\end{table}

\begin{table}[htb!]
  \centering
  \renewcommand\arraystretch{1}
  \caption{Detection Accuracy of Adversarial Images for Proposed Method}\label{2}
  \begin{tabular}{c|c|c|c|c|c|c}
  \hlinew{1pt}
  size &method&$\epsilon=1$ &$\epsilon=2$ &$\epsilon=3$ &$\epsilon=4$&$\epsilon=5$ \\
  \hlinew{1pt}
  $64\times64$&FGSM &93.6\%&94.5\% &96.1\% &96.3\%  &95.9\%\\
              &TGSM &90.0\%&91.6\% &93.0\% &93.3\%  &93.0\%\\
  \hlinew{0.5pt}
  $128\times128$&FGSM &95.9\%&97.1\% &97.8\% &98.4\%  &98.7\%\\
                &TGSM &94.5\%&95.6\% &96.2\% &96.8\%  &97.1\%\\
  \hlinew{0.5pt}
  $256\times256$&FGSM &97.4\%&98.5\% &98.8\% &98.9\%  &98.6\%\\
                &TGSM &94.3\%&95.1\% &95.3\% &95.4\%  &95.2\%\\
  \hlinew{0.5pt}
  $512\times512$&FGSM &99.2\%&99.3\% &99.3\% &99.4\%  &99.3\%\\
                &TGSM &98.1\%&98.2\% &98.3\% &98.3\%  &98.2\%\\
  \hlinew{1pt}
  \end{tabular}
\end{table}

\section{Conclusions}

Although lots of forensic methods based on deep learning have achieved state-of-the-art performance, there is still a defect that they are vulnerable to adversarial examples. To against this potential threat, we propose a secure deep learning forensic method in which a penalty term is added to the loss function and  both normal images and adversarial images are fused into training set. The effectiveness of proposed scheme are evaluated by a set of experiments using four different sizes of images and two different methods of generating adversarial examples with a series of noise strength. Experimental results show that proposed scheme is effective to resist against the attack of adversarial examples.

\section*{Acknowledgments}

This work was supported in part by National NSF of China (61672090, 61332012), the National Key Research and Development of China (2016YFB0800404), Fundamental Research Funds for the Central Universities (2015JBZ002, 2017YJS054). We greatly acknowledge the support of NVIDIA Corporation with the donation of the GPU used for this research.

\newpage
%\bibliographystyle{ieicetr}% bib style
%\bibliography{}% your bib database

\newpage

%{NRR.jpg}}

\profile[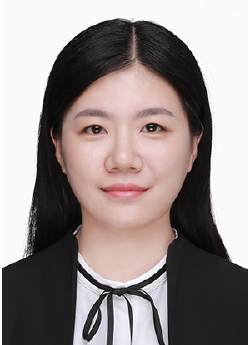] {Wei Zhao}{received the B.S. degree from Yanshan University, Qinhuangdao, China, in 2016. She is currently a master of the Institute of Information Science, Beijing Jiaotong University, Beijing, China. Her current research interests include image precessing, digtial image forensics, pattern recognition, and deep learning.}

\profile[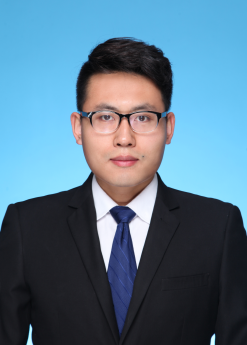]{Pengpeng Yang}{received the B.S. degree from Ludong University, Yantai, China, in 2014. He is currently a  PhD candidate of the Institute of Information Science, Beijing Jiaotong University, Beijing, China. His current research interests include image precessing, digtial image forensics, pattern recognition, and deep learning. }

\profile[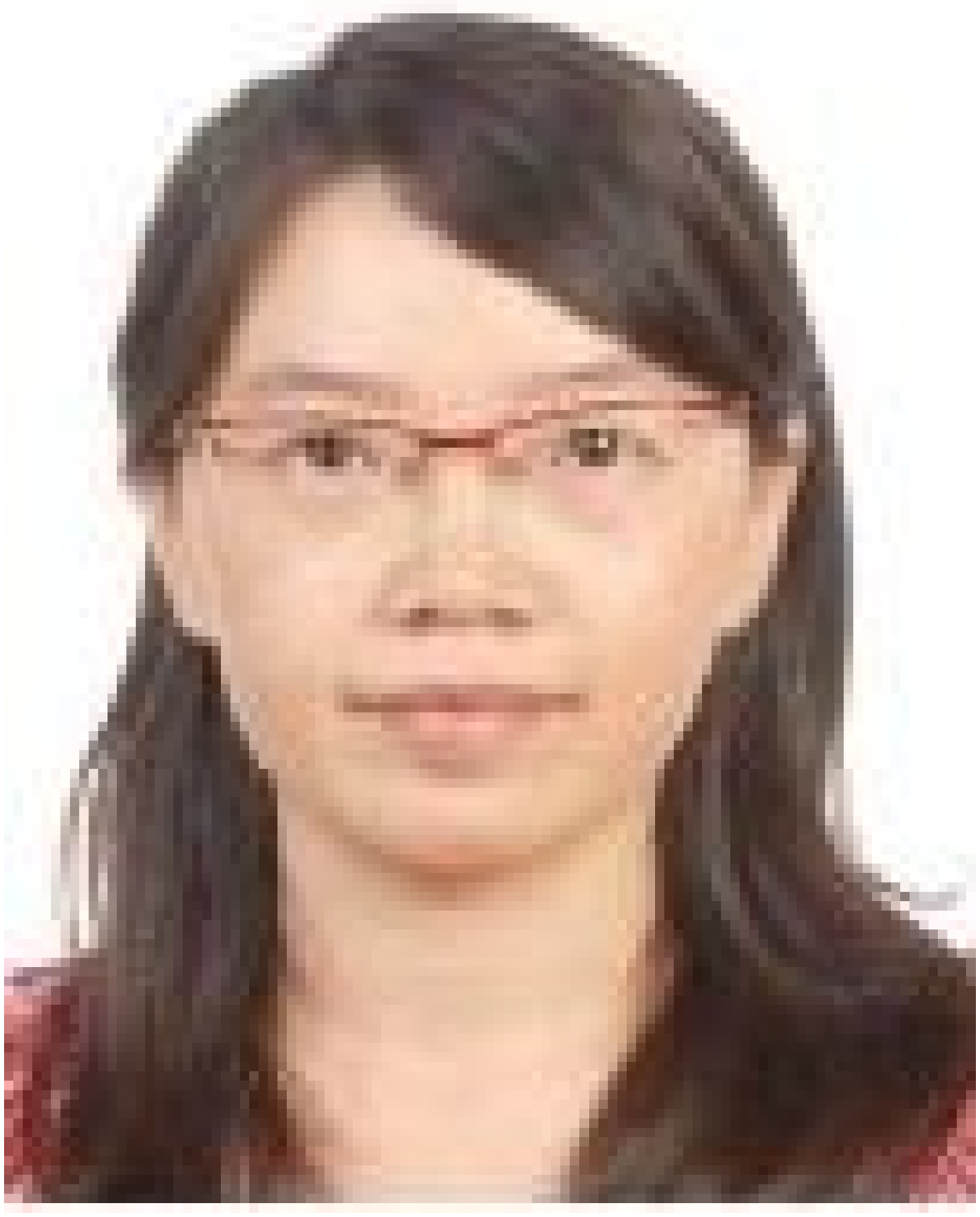]{Rongrong Ni}{received the B.S. degree and the Ph.D. degree from Beijing Jiaotong University (BJTU), Beijing, China, in 1998, and 2005, respectively. Since 2005, she has been the faculty of the School of Computer and Information Technology and the Institute of Information Science, BJTU, where she is a Professor since 2013. Her current research interests include image processing, data hiding and digital forensics, pattern recognition, and computer vision. She was selected in the Beijing Science and Technology Stars Projects in 2008, and was awarded the Jeme Tien Yow Special Prize in Science and Technology in 2009. She is the Principal Investigator of three projects funded by the Natural Science Foundation of China. She has participated in the 973 Program, the 863 Program, and international projects. She has published more than 80 papers in academic journals and conferences, and holds six national patents.}

\profile[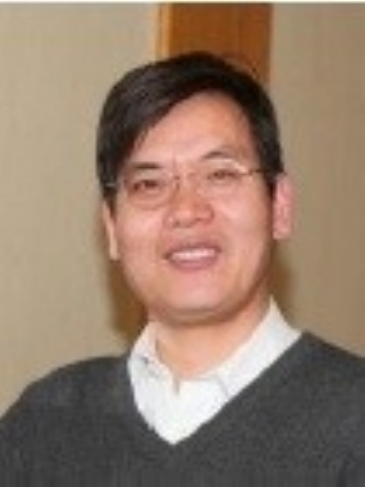]{Yao Zhao}{received the B.S. degree in radio engineering from Fuzhou University, Fuzhou, China, in 1989, the M.E. degree in radio engineering from Southeast University, Nanjing, China, in 1992, and the Ph.D. degree from the Institute of Information Science, Beijing Jiaotong University (BJTU), Beijing, China, in 1996. He was an Associate Professor at BJTU in 1998, where he became a Professor in 2001. He was a Senior Research Fellow with the Information and Communication Theory Group, Faculty of Information Technology and Systems, Delft University of Technology, Delft, The Netherlands, from 2001 to 2002. He is currently the Director with the Institute of Information Science, BJTU. His current research interests include image/video coding, digital watermarking and forensics, and video analysis and understanding. He is currently leading several national research projects from the 973 Program, 863 Program, and the National Science Foundation of China. He serves on the editorial boards of several international journals, including as an Associate Editor of the IEEE TRANSACTIONS ON CYBERNETICS, Associate Editor of the IEEE SIGNAL PROCESSING LETTERS, and Area Editor of Signal Processing: Image Communication. Dr. Zhao was a recipient of the Distinguished Young Scholar by the National Science Foundation of China in 2010 and Chang Jiang Scholar of the Ministry of Education of China in 2013.}

\profile[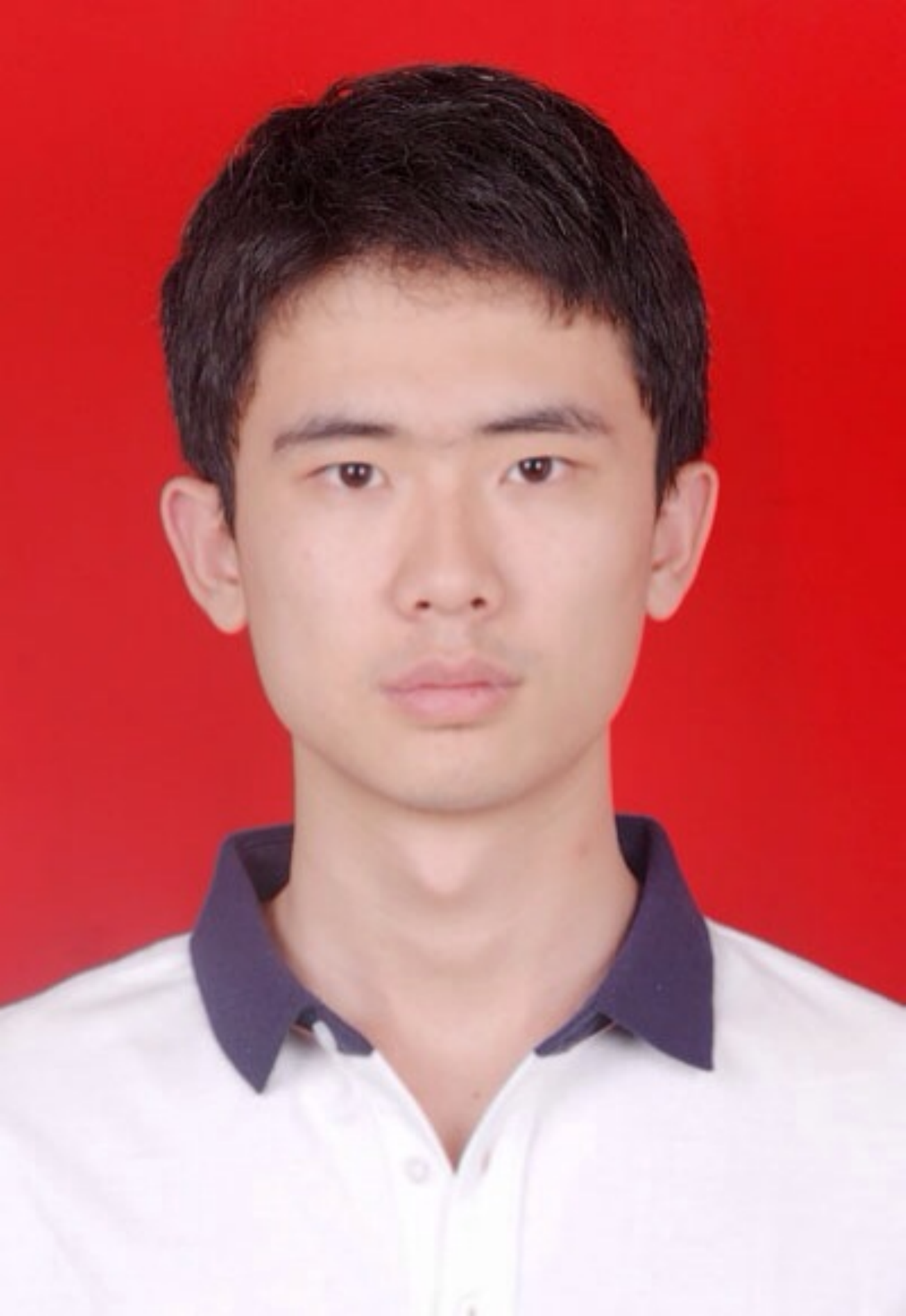]{Haorui Wu}{received the B.S. degree from Xi'an University of Posts and Telecommunications, Xi'an, China, in 2016. He is currently a PhD candidate of the Institute of Information Science, Beijing Jiaotong University, Beijing, China. His current research interests include image precessing, digtial image forensics, pattern recognition, and deep learning.}

\end{document}